\documentclass[runningheads,a4paper,hidelinks]{llncs}
\usepackage[T1]{fontenc}
\usepackage{graphicx}
\usepackage{amsmath}
\usepackage{amsfonts} 
\usepackage{amssymb} 
\usepackage{algorithm} 
\usepackage{algpseudocode}

\usepackage[utf8]{inputenc}
\usepackage{lmodern}
\usepackage{tikz}

\usetikzlibrary{arrows.meta,positioning,fit,calc}
\usepackage{adjustbox} 
\usepackage{microtype}
\usepackage{setspace}
\usepackage{tabularx}    
\usepackage{array}       
\usepackage{booktabs}     
\usepackage{amsmath,amssymb,amsfonts,mathtools}
\usepackage{tabularx} 
\usepackage{graphicx}
\usepackage{booktabs}     
\usepackage{multirow}
\usepackage{makecell}
\usepackage{siunitx}
\usepackage{subcaption}
\usepackage{caption}
\usepackage{float}
\usepackage[skins,breakable]{tcolorbox}
\usepackage{xcolor}
\usepackage{hyperref}
\usepackage{amsmath,amssymb,amsfonts} 
\usepackage{algpseudocode} 
\usepackage{tikz}
\usetikzlibrary{arrows.meta,positioning,fit,calc,shapes.geometric}
\usepackage{marvosym}
\usepackage{enumitem}
\setlist{nosep,leftmargin=*}
\usepackage{wrapfig}
\usepackage{graphicx}
\begin{document}
\title{Dodgersort: Uncertainty-Aware VLM-Guided Human-in-the-Loop Pairwise Ranking}
\titlerunning{VLM-Guided HIL Pairwise Ranking}
%
\author{Yujin Park\inst{1}\orcidID{0009-0001-8988-5698} \and
Haejun Chung\inst{1}\orcidID{0000-0001-8959-237X} \and
Ikbeom Jang\inst{2}\textsuperscript{(\Letter)}\orcidID{0000-0002-6901-983X}}
\authorrunning{Y. Park et al.}
%
\institute{Hanyang University, Seoul, Republic of Korea \\
\email{\{yujin1019a,haejun\}@hanyang.ac.kr} \and
Hankuk University of Foreign Studies, Yongin, Republic of Korea \\
\email{ijang@hufs.ac.kr}}

\maketitle             

\begin{abstract}
Pairwise comparison labeling is emerging as it yields higher inter-rater reliability than conventional classification labeling, but exhaustive comparisons require quadratic cost. We propose Dodgersort, which leverages CLIP-based hierarchical pre-ordering, a neural ranking head and probabilistic ensemble (Elo, BTL, GP), epistemic--aleatoric uncertainty decomposition, and information-theoretic pair selection. It reduces human comparisons while improving the reliability of the rankings. In visual ranking tasks in medical imaging, historical dating, and aesthetics, Dodgersort achieves a 11--16\% annotation reduction while improving inter-rater reliability. Cross-domain ablations across four datasets show that neural adaptation and ensemble uncertainty are key to this gain. In FG-NET with ground-truth ages, the framework extracts 5--20$\times$ more ranking information per comparison than baselines, yielding Pareto-optimal accuracy--efficiency trade-offs.


\end{abstract}

\section{Introduction}

Pairwise comparison is a widely accepted method for subjective annotation across many domains, including vision, language, and broader data-driven applications. In this work, we focus on visual ranking tasks such as medical imaging, historical curation, and aesthetics assessment, where expert consistency directly impacts the quality of downstream decisions. Compared to absolute ratings, pairwise judgments generally yield higher inter-rater reliability and more stable supervision~\cite{kalpathy2016plus}. However, the quadratic cost of exhaustive pairwise annotation ($O (n^2)$) severely limits scalability: ranking 1,000 images requires nearly 500{,}000 comparisons, making exhaustive annotation impractical for large-scale applications.

Sorting-based active learning methods alleviate this burden by reducing complexity to $O (n\log n)$, querying informative pairs during the sorting process~\cite{jang2022decreasing}. In parallel, pre-trained vision--language models (VLMs) such as CLIP~\cite{radford2021learning} provide zero-shot semantic priors that can coarsely order images from a text prompt describing the ranking criterion. This suggests a promising direction: use VLM-based pre-ordering to eliminate many trivial comparisons and reserve human effort for uncertain cases.

Despite this progress, three challenges remain. \textbf{ (1)} Zero-shot VLM rankings are noisy at fine granularity: CLIP separates broad categories (e.g., infants vs.\ elderly) but often misorders items within a local neighborhood (e.g., similar-age faces), so VLM scores alone are insufficient for high-quality rankings. \textbf{ (2)} Existing learning mechanisms are not tailored to refine such noisy VLM priors from a small number of pairwise labels into a calibrated, sample-efficient ranking model that generalizes to unseen pairs while tracking its own confidence. \textbf{ (3)} Even with such a model, we still need a principled strategy to decide \emph{which} pairs to query and \emph{when} to trust automation, balancing exploration (epistemic uncertainty) and exploitation while accounting for task-inherent ambiguity (aleatoric uncertainty) that should trigger human judgment rather than automatic decisions.

We propose \textbf{Dodgersort}, a reliable, VLM-assisted framework for ranking annotation, which addresses these challenges through three key innovations. First, we integrate CLIP-based hierarchical pre-ordering with a lightweight text-conditioned neural ranking head and probabilistic ensemble (GP, Elo, BTL) to refine noisy VLM priors from minimal human feedback. Second, we decompose epistemic and aleatoric uncertainty to enable principled automated decision-making when selecting maximally informative pairs by balancing information gain, model disagreement, and novelty. Third, our experiments demonstrate Pareto-optimal accuracy--efficiency trade-offs: on human-annotated datasets, we achieve equal or better inter-rater reliability with 13\% fewer comparisons, while on simulated ground truth we convert each additional comparison into ranking accuracy 5--20$\times$ more efficiently than random labeling.

\section{Related Work}

\textbf{Pairwise annotation and scalability.}
Pairwise comparison, formalized by Bradley--Terry~\cite{bradley1952rank}, provides reliable supervision for subjective tasks but incurs quadratic annotation cost ($O (n^2)$). Sorting-based approaches~\cite{jang2022decreasing} reduce this to $O (n\log n)$ by integrating active learning into MergeSort, querying human annotators only for pairs needed to complete the sort. More recent work leverages machine learning to guide sorting~\cite{bai2023sorting}, but still requires human input for every requested comparison and lacks principled strategies to automate confident decisions. We build on this line of work by combining VLM-based pre-ordering with ensemble-based refinement so that many comparisons can be handled automatically while preserving ranking quality.

\textbf{Active learning for ranking.}
Active preference learning aims to minimize annotation costs by querying informative pairs, typically via information gain~\cite{saar2004active} or model disagreement~\cite{houlsby2011bayesian}. GURO~\cite{bergstrom2024active} combines multiple uncertainty signals for robust selection. Such methods primarily target epistemic uncertainty, however, and do not explicitly distinguish between task-inherent aleatoric uncertainty~\cite{kendall2019geometry}
, which can lead to over-automation on ambiguous pairs. They also usually assume no strong prior and start from scratch. In contrast, Dodgersort uses the epistemic--aleatoric decomposition not merely to prioritize which pairs to query, but to enable a principled binary decision: pairs with low epistemic uncertainty are automated regardless of aleatoric ambiguity, while pairs with high task-inherent ambiguity trigger human judgment rather than being silently automated---a distinction absent in GURO, where every selected pair still requires human input. Combined with VLM-based pre-ordering to eliminate trivial comparisons from the outset, this yields a genuinely semi-automatic pipeline rather than a purely query-prioritization scheme.

\textbf{Vision--language models for ranking.}
Large-scale VLMs such as CLIP~\cite{radford2021learning} and its variants~\cite{shen2022k} 
support zero-shot image ordering from text prompts, with prompt learning~\cite{zhou2022conditional} 
improving domain adaptation. EZ-Sort~\cite{park2025ez} combines CLIP-based hierarchical pre-ordering with an uncertainty-aware MergeSort and achieves substantial annotation reduction on small datasets. In our setting, EZ-Sort is limited by heuristic uncertainty scores that conflate model confidence and task difficulty, fixed Elo updates that do not learn richer ranking patterns, and performance degradation at larger scales ($n \geq 100$). Dodgersort builds on EZ-Sort by replacing heuristic scores with a probabilistic ensemble plus a lightweight neural ranking head on frozen CLIP features, and by introducing an uncertainty-aware pair selection and automation policy. This yields a semi-automatic VLM-guided ranking pipeline that scales more effectively while maintaining high ranking quality. We note the framework is compatible with other VLM backbones beyond CLIP.

\begin{figure}
    \centering
    \includegraphics[width=1\linewidth]{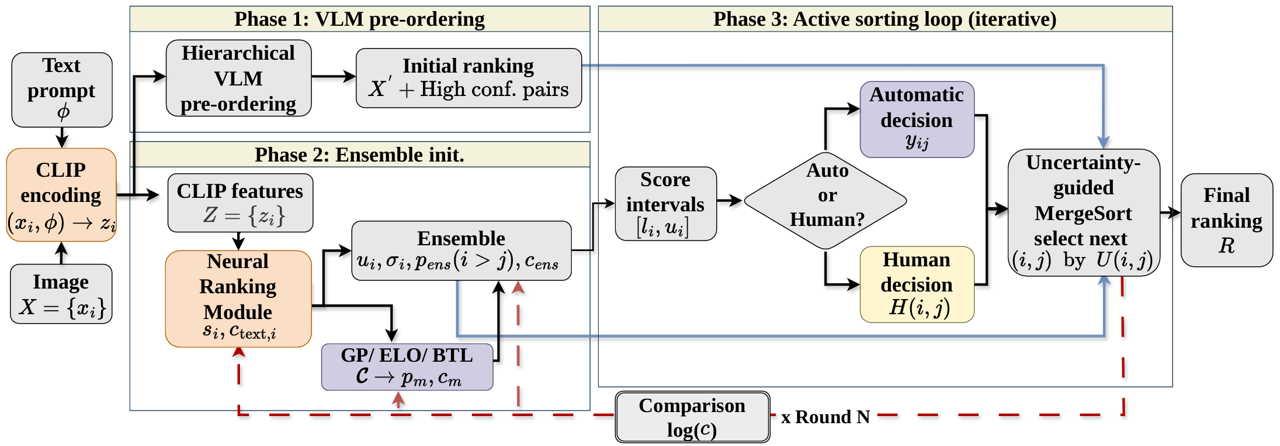}

\caption{Pipeline of Dodgersort. Phase~1: VLM hierarchical pre-ordering produces coarse ranking. Phase~2: Initialize neural ranking head and probabilistic ensemble (GP, Elo, BTL). Phase~3: Uncertainty-guided MergeSort loop automates confident pairs or queries human, updating ensemble with feedback to produce final ranking $\mathcal{R}$.}
\label{fig:pipeline}
\end{figure}

\section{Methods}

\subsection{Problem Formulation}

Given $n$ images $\mathcal{X}$ and a ranking criterion expressed as a text prompt $\phi$, we aim to learn a ranking function $f: \mathcal{X} \to \mathbb{R}$ from a minimal set of human pairwise comparisons $\mathcal{D} = \{(v \succ u)\}$. The goal is to minimize queries $|\mathcal{D}|$ while maximizing ranking quality (e.g., Kendall's $\tau$).

\subsection{Hierarchical VLM Pre-Ordering}
We leverage CLIP~\cite{radford2021learning} to establish an initial coarse ordering without human input. Given a ranking theme $\phi$, we construct $B$ hierarchical text prompts $\{t_b\}_{b=1}^B$ describing different quality levels (e.g., for age: ``infant'', ``teenager'', ``adult'', ``elderly''). Each image receives a soft bin assignment via temperature-scaled softmax: $p_{ib} = \text{softmax} (s_{ib}/T)$, where $T=2.0$ is a temperature hyperparameter and $s_{ib}$ is the CLIP cosine similarity between image $x_i$ and prompt $t_b$. The expected bin score $\mathbb{E}[\text{bin}_i] = \sum_{b=1}^B b \cdot p_{ib}$ provides the initial ordering, with confidence $c_i^{\text{pre}} = \min (0.75, \max_b p_{ib})$. For images with large bin gaps ($|\text{bin} (u) - \text{bin} (v)| \geq 2$) and high confidence ($\min (c_u, c_v) \geq 0.65$), we seed high-confidence edges with weight $0.74$ into the comparison graph.
These high-confidence pairs form the initial comparison dataset $\mathcal{D}_0$, which initializes probabilistic models: \textbf{Elo ratings} are set proportional to bin assignments following EZ-Sort~\cite{park2025ez} ($R_i^{ (0)} = 1200 + 150 \cdot \text{bin} (i) + \epsilon$, where $\epsilon \sim \mathcal{U} (-75, 75)$ and bins $\in \{0,\ldots,4\}$), and \textbf{neural ranking head} undergoes initial training on $\mathcal{D}_0$ before active learning begins.

\begin{figure}
    \centering
    \includegraphics[width=1.0\linewidth]{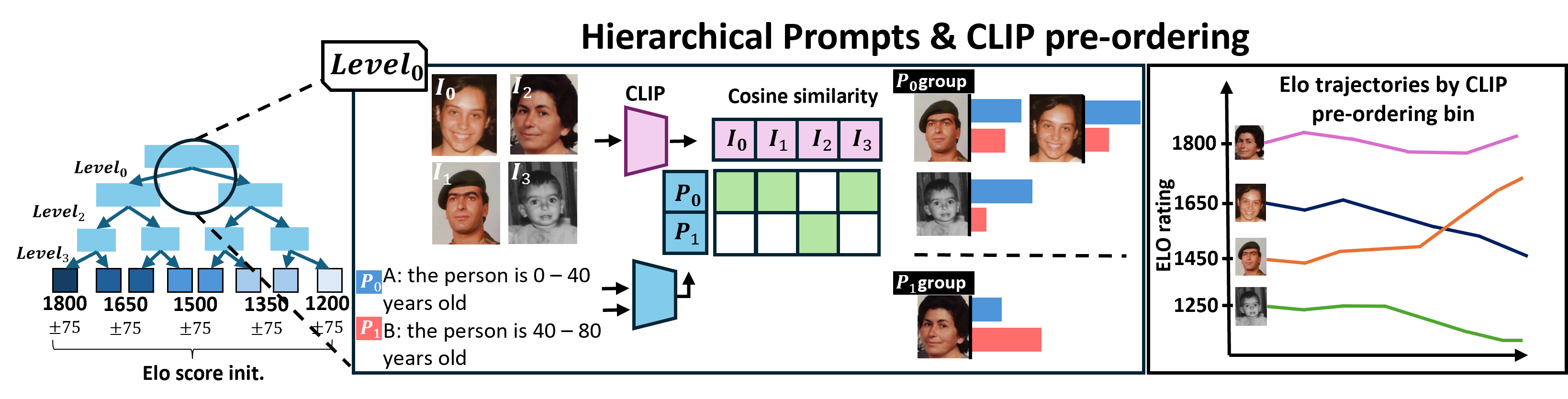}
\caption{Hierarchical VLM pre-ordering. \textbf{Left:} Prompt structure with $B$ bins. \textbf{Center:} CLIP assigns soft bin probabilities $\{p_{ib}\}$ via image-text cosine similarities. \textbf{Right:} Elo rating trajectories show that proper bin initialization effectively accelerates convergence.}
\label{fig:hierarchical_preordering}
\end{figure}

\subsection{Neural Ranking Module}

\textbf{Architecture.}
The ranking head refines frozen CLIP features through $L=2$ cross-attention 
layers (512-dim) fusing image-text representations, with dual outputs: regression 
scores $s_i = \sigma (W_r \mathbf{h}_i) \in [0,1]$ and pairwise logits 
$\ell_{ij} = W_p \cdot \text{RelAttn} (\mathbf{h}_i, \mathbf{h}_j, \mathcal{R})$ using 
learnable relation tokens ($\mathcal{R} \in \mathbb{R}^{16 \times 512}$), totaling 
1.3M trainable parameters. For ordinal tasks, we optionally project to Poincaré ball 
$\mathcal{B}_c^{128}$ ($c=0.05$).

\textbf{Training.}
The multi-task loss combines regression and ranking:
\[
\mathcal{L} = \tfrac{1}{2}\text{BCE} (s_i, y) + \tfrac{1}{2}\text{BCE} (s_j, 1-y) + 
\tfrac{1}{2}\text{BCE}_{\text{logit}} (\ell_{ij}, y) + 10^{-3}\mathcal{L}_{\text{reg}},
\].
Updates occur every 50 comparisons. Training is performed \emph{asynchronously} in a background thread so the frozen previous model serves queries during retraining, ensuring zero annotator idle time. Elo and BTL update instantaneously ($O(1)$) after every comparison, so ensemble decisions remain responsive even during neural head retraining.
\textbf{Hyperbolic projection (optional).}
For strongly ordinal datasets, we optionally use a compact hyperbolic projection~\cite{nickel2017poincare} to better preserve order structure; otherwise this module is disabled with no change to the rest of the pipeline.

\subsection{Gaussian Process Preference Learning}

Following~\cite{chu2005preference}, we model latent ranking scores as draws from a Gaussian process (GP), $\mathbf{f} \sim \mathcal{GP} (\mathbf{0}, \mathbf{K})$, where the covariance is given by an ARD (Automatic Relevance Determination) kernel with per-dimension lengthscales $\{\ell_d\}_{d=1}^D$ and signal variance $\sigma_f^2=1.0$: 
\[
K (x_i, x_j) = \sigma_f^2 \exp\left (-\frac{1}{2}\sum_{d=1}^D \frac{ (x_i^d - x_j^d)^2}{\ell_d^2}\right),
\]
where $D$ is the CLIP feature dimension (512 for ViT-B/32) and $\ell_d=1.0$ for all dimensions initially. For preference $v \succ u$, the probit likelihood is $P (v \succ u \mid \mathbf{f}) = \Phi ( (f_v - f_u)/\sqrt{2}\sigma)$ with noise scale $\sigma=0.5$, where $\Phi$ is the standard normal CDF. We obtain the MAP estimate $\mathbf{f}_{\text{MAP}}$ by minimizing
\[
S (\mathbf{f}) = -\sum_k \log \Phi (z_k) + \frac{1}{2}\mathbf{f}^\top \mathbf{K}^{-1} \mathbf{f}
\]
via L-BFGS-B (max 100 iterations). The Laplace approximation yields posterior covariance $\boldsymbol{\Sigma}_{\text{post}} = (\mathbf{K}^{-1} + \boldsymbol{\Lambda})^{-1}$, enabling uncertainty quantification. For pair $ (i,j)$, the predictive probability is $P (i \succ j \mid \mathcal{D}) \approx \Phi ( (\mu_i - \mu_j)/\sigma')$, where $\sigma'^2 = 2\sigma^2 + \Sigma_{ii} + \Sigma_{jj} - 2\Sigma_{ij}$ combines epistemic (posterior variance) and aleatoric (noise) uncertainty.

\textbf{Practical approximation.}
To avoid the cubic time and quadratic memory overhead of exact GP inference, we restrict the GP head to an active subset of $M$ items and $M_{\text{pairs}}$ subsampled comparisons. We directly optimize latent scores $\mathbf{f}\in\mathbb{R}^M$ for the probit objective with $\ell_2$ regularization using first-order methods, ensuring efficient and accurate probability estimates.
\paragraph{Large-$n$ mode and theoretical justification.}
For $n > 300$, exact GP inference ($O(n^3)$) causes unacceptable latency, so Dodgersort disables it ($w_{\text{gp}}{=}0$, omitting $I_{\text{gain}}$). This omission is theoretically justified by the diminishing marginal utility of GP at scale: as the $\Omega(n \log n)$ comparison graph densifies, the GP posterior is rapidly dominated by empirical likelihood. Consequently, explicit GP information gain ($I_{\text{gain}}$) becomes highly redundant with the empirical model disagreement ($u_{\text{dis}}$) already captured by the scalable 3-model ensemble (text, Elo, BTL). Thus, omitting GP at larger scales preserves real-time responsiveness and sample efficiency with negligible ranking degradation.

\subsection{Ensemble-Based Automation}

We integrate four complementary models with adaptive weights $\{w_m\}_{m=1}^4$: neural ranking head producing $p_{\text{text}} (i \succ j)$ with confidence $c_{\text{text}} = |s_i - s_j| \cdot (1 - \text{training\_loss})$; Elo ratings with adaptive K-factor (initial $K=128$, decaying to 64 after 100 comparisons) yielding scores $R_i$ and uncertainty $\sigma_i = 2K/\sqrt{1 + n_i}$; BTL model providing strengths $\{\pi_i\}$ via MM algorithm (10 iterations, regularization $\lambda=1.0$) and probability $p_{\text{BTL}} (i \succ j) = \pi_i/ (\pi_i + \pi_j)$; and Gaussian Process with $p_{\text{GP}} (i \succ j)$ and uncertainty score $c_{\text{GP}}$ from the posterior. 

Initial ensemble weights are set based on validation performance: without GP, $w_{\text{text}}=0.4$, $w_{\text{elo}}=0.3$, $w_{\text{btl}}=0.3$; with GP enabled, $w_{\text{text}}=0.3$, $w_{\text{elo}}=0.25$, $w_{\text{btl}}=0.25$, $w_{\text{gp}}=0.2$. These remain fixed during active learning to maintain computational efficiency. The ensemble prediction is
\[
p_{\text{ens}} (i \succ j) = \frac{\sum_m w_m c_m \, p_m (i \succ j)}{\sum_m w_m c_m},
\]
with overall confidence $c_{\text{ens}} = \min (0.95, (|p_{\text{ens}} - 0.5| + \bar{c})/2)$, where $\bar{c}$ is average model confidence.

\textbf{Automation decision.}
For each pair $ (i,j)$ requested by MergeSort, we automatically resolve it if any of four conditions hold: (1)interval certification: score or Elo confidence intervals are non-overlapping ($[\mu_i - \gamma\sigma_i, \mu_i + \gamma\sigma_i] \cap [\mu_j - \gamma\sigma_j, \mu_j + \gamma\sigma_j] = \emptyset$ where $\gamma=2.0$); (2)GP confidence: $c_{\text{GP}} \geq 0.75$ with epistemic uncertainty $< 0.25$; (3)ensemble agreement: $c_{\text{ens}} \geq \theta_{\text{base}}$ with $\geq$3 models agreeing, where $\theta_{\text{base}} = 0.65 + 0.10 \cdot \min (1, n/200)$ adapts to dataset size. For top-10\% items, thresholds increase by 0.05--0.10 to prevent high-impact errors.

\subsection{Information-Theoretic Pair Selection}

For pairs requiring human judgment, we select the most informative using GURO-inspired composite utility~\cite{bergstrom2024active}:
\[
U (i,j) = \lambda_{\text{epi}} \, u_{\text{epi}} + \lambda_{\text{ale}} \, u_{\text{ale}} + \lambda_{\text{gain}} \, I_{\text{gain}} + \lambda_{\text{dis}} \, u_{\text{dis}} + \lambda_{\text{nov}} \cdot \text{novelty},
\]

where $u_{\text{epi}}$ is weighted ensemble variance (epistemic uncertainty), $u_{\text{ale}} = H (p_{\text{ens}})$ is prediction entropy (aleatoric uncertainty), $I_{\text{gain}} = \Sigma_{ii} + \Sigma_{jj} - 2\Sigma_{ij}$ is GP information gain, $u_{\text{dis}}$ measures model disagreement, and novelty penalizes recently compared pairs. 

\begin{wraptable}{r}{0.47\textwidth} 
\vspace{-4mm}
\centering
\caption{Human annotation count averaged across EyePACS, DHCI, TAD66k.}
\label{tab:comparison_count_fgnet}
\resizebox{0.95\linewidth}{!}{
\begin{tabular}{lcccc}
\toprule
\textbf{Size} & \textbf{Exhaustive} & \textbf{Sort} & \textbf{EZ-Sort} & \textbf{Ours}\\
\midrule
$n{=}30$  & 435   & 126 & 90  & \textbf{80} \\
$n{=}50$  & 1{,}225 & 240 & \textbf{144} & 147\\
$n{=}100$ & 4{,}950 & 582 & 475 & \textbf{400}\\
\bottomrule
\end{tabular}
} 
\end{wraptable}
We set $\lambda_{\text{epi}}=0.5$, $\lambda_{\text{ale}}=0.4$, $\lambda_{\text{gain}}=0.3$, $\lambda_{\text{dis}}=0.15$, $\lambda_{\text{nov}}=0.1$ based on validation experiments, balancing exploration (epistemic, information gain) and exploitation (aleatoric, disagreement) while avoiding redundant queries.

\textbf{Complexity.} Dodgersort requires $O(n \log n)$ total comparisons. Although the system introduces computational overhead via VLM pre-ordering ($O(nB)$) and periodic ensemble refitting, we mitigate latency by restricting the computationally expensive GP to small subsets ($n \le 300$) and relying on scalable models (Elo, BTL) for larger sets. Ultimately, the system's efficiency is defined by \emph{human effort reduction}. Dodgersort automates 30--40\% of the queries, requiring only $0.6$--$0.7 \cdot n\log n$ human comparisons. Since human judgments (5--30 seconds/pair) strictly bottleneck the annotation process, the time saved by reducing human queries far outweighs the model's computational overhead.

\begin{table}[t]
\caption{Inter-rater reliability across datasets. Reported: \textit{Sp} (Spearman), \textit{Ke} (Kendall), \textit{Pe} (Pearson), \textit{ICC} (intraclass correlation). The last three methods (Sort comparison, EZ-Sort, and Dodger-Sort) are pairwise comparison-based approaches.}
\centering
\scriptsize
\resizebox{\linewidth}{!}{%
\begin{tabular}{lcccccccccccc}
\toprule
 & \multicolumn{4}{c}{\textbf{Retina (EyePACS)}} &
   \multicolumn{4}{c}{\textbf{Historical (DHCI)}} &
   \multicolumn{4}{c}{\textbf{Aesthetics (TAD66k)}} \\
\cmidrule (lr){2-5}\cmidrule (lr){6-9}\cmidrule (lr){10-13}
\textbf{Method} & Sp & Ke & Pe & ICC & Sp & Ke & Pe & ICC & Sp & Ke & Pe & ICC \\
\midrule
Classification &
\makecell{0.53\\\scriptsize ($\pm$0.06)} &
\makecell{0.46\\\scriptsize ($\pm$0.07)} &
\makecell{0.54\\\scriptsize ($\pm$0.07)} &
\makecell{0.75\\\scriptsize (N/A)} &
\makecell{0.39\\\scriptsize ($\pm$0.06)} &
\makecell{0.33\\\scriptsize ($\pm$0.05)} &
\makecell{0.42\\\scriptsize ($\pm$0.06)} &
\makecell{0.68\\\scriptsize (N/A)} &
\makecell{0.46\\\scriptsize ($\pm$0.11)} &
\makecell{\textbf{0.38}\\\scriptsize ($\pm$0.12)} &
\makecell{0.50\\\scriptsize ($\pm$0.16)} &
\makecell{0.64\\\scriptsize (N/A)} \\
\addlinespace[1pt]
Sort comparison~\cite{jang2022decreasing} &
\makecell{0.72\\\scriptsize ($\pm$0.07)} &
\makecell{0.56\\\scriptsize ($\pm$0.06)} &
\makecell{0.72\\\scriptsize ($\pm$0.07)} &
\makecell{0.89\\\scriptsize (N/A)} &
\makecell{0.47\\\scriptsize ($\pm$0.17)} &
\makecell{0.35\\\scriptsize ($\pm$0.15)} &
\makecell{0.47\\\scriptsize ($\pm$0.17)} &
\makecell{0.78\\\scriptsize (N/A)} &
\makecell{0.43\\\scriptsize ($\pm$0.01)} &
\makecell{0.34\\\scriptsize ($\pm$0.01)} &
\makecell{0.43\\\scriptsize ($\pm$0.01)} &
\makecell{0.60\\\scriptsize (N/A)} \\
\addlinespace[1pt]
EZ-Sort~\cite{park2025ez} &
\makecell{0.85\\\scriptsize ($\pm$0.09)} &
\makecell{\textbf{0.76}\\\scriptsize ($\pm$0.14)} &
\makecell{0.85\\\scriptsize ($\pm$0.09)} &
\makecell{0.94\\\scriptsize (N/A)} &
\makecell{0.47\\\scriptsize ($\pm$0.17)} &
\makecell{0.39\\\scriptsize ($\pm$0.15)} &
\makecell{0.47\\\scriptsize ($\pm$0.16)} &
\makecell{0.73\\\scriptsize (N/A)} &
\makecell{0.46\\\scriptsize ($\pm$0.02)} &
\makecell{0.31\\\scriptsize ($\pm$0.02)} &
\makecell{0.56\\\scriptsize ($\pm$0.06)} &
\makecell{0.71\\\scriptsize (N/A)} \\
\addlinespace[1pt]
Dodger-Sort (ours) &
\makecell{\textbf{0.86}\\\scriptsize ($\pm$0.07)} & 
\makecell{0.67\\\scriptsize ($\pm$0.06)} & 
\makecell{\textbf{0.86}\\\scriptsize ($\pm$0.06)} & 
\makecell{\textbf{0.95}\\ \scriptsize (N/A)} &
\makecell{\textbf{0.60}\\\scriptsize ($\pm$0.05)} & 
\makecell{\textbf{0.43}\\\scriptsize ($\pm$0.04)} & 
\makecell{\textbf{0.60}\\\scriptsize ($\pm$0.05)} & 
\makecell{\textbf{0.82}\\ \scriptsize (N/A)} &
\makecell{\textbf{0.47}\\\scriptsize ($\pm$0.10)} & 
\makecell{0.32\\\scriptsize ($\pm$0.07)} & 
\makecell{\textbf{0.57}\\\scriptsize ($\pm$0.12)} & 
\makecell{\textbf{0.72}\\\scriptsize (N/A)} \\
\bottomrule
\end{tabular}}
\label{tab:inter-rater}
\end{table}

\section{Experiments and Results}

\subsection{Setup and Metrics}

\textbf{Datasets.} 
We evaluate on four visual ranking datasets: \small\textbf{Retina (EyePACS)} --diabetic retinopathy assessment (subset of 28,792 images); \small\textbf{Historical (DHCI)} --temporal photo ordering (subset of 450 images, 1930s--1970s); \small\textbf{Aesthetics (TAD66k)} --visual appeal ranking (subset of 66,327 images); \small\textbf{Face-Age (FG-NET)} --age-ordered faces used for controlled experiments with dataset sizes ranging from 30 to 300 images to evaluate scaling behavior.

\textbf{Annotation protocol and dataset selection.}
We evaluate Dodgersort through two complementary experimental paradigms:

\textbf{Inter-rater reliability studies} (Table~\ref{tab:inter-rater}) assess whether the ranking framework maintains high annotator agreement under realistic noise conditions. Five annotators with computer vision expertise provided independent pairwise rankings on Retina, Historical, and Aesthetics after training sessions. Face-Age was excluded due to ceiling effects: prior work~\cite{park2025ez} showed all methods achieve ICC $\geq$ 0.97 on age ranking, offering limited discriminative power. We report Spearman, Kendall, Pearson correlations and ICC(2,k) using a two-way random-effects model. Specifically, each annotator's final item ranking (derived from their pairwise comparisons via the same scoring pipeline) serves as one rater column; ICC(2,k) is then computed across the five raters' ranking vectors to quantify cross-annotator consistency.

\textbf{Controlled experiments} measure algorithmic performance using numeric labels as ground truth, enabling objective accuracy measurement (Kendall's $\tau$) without human annotation costs. For \textbf{component ablation} (Table~\ref{tab:cross_domain_ablation_vertical}), we use four datasets—Retina, Face-Age, Historical, and Aesthetics—to validate that each component contributes robustly across diverse ranking criteria spanning medical diagnosis (3-class severity), age estimation (continuous), temporal ordering (5-decade bins), and aesthetics assessment (aggregated ratings). We performed \textbf{accuracy-to-efficiency comparison with EZ-Sort} (Table~\ref{tab:tau_vs_labels}) using Face-Age. This dataset provides objective, fine-grained continuous chronological ages, avoiding tied rankings and thus enabling clean algorithmic comparison.

\textbf{Baselines.}
We compare against: (1) \emph{Classification} trained on absolute scores, (2) \emph{Sort comparison}~\cite{jang2022decreasing}, and (3) \emph{EZ-Sort}~\cite{park2025ez} with CLIP hierarchical pre-ordering. 
\textbf{Evaluation metrics.}
Ranking quality: Spearman ($\rho$), Kendall ($\tau$), Pearson ($r$), ICC(2,k). Efficiency: $\Delta\text{HC}$ denotes the reduction in human comparisons relative to baseline (positive = fewer comparisons needed). Statistical significance via Wilcoxon signed-rank test ($p<0.05$). 
\textbf{Implementation.}
CLIP ViT-B/32 (frozen), CPU-trained ranking head, ensemble weights via grid search on 20\% validation split. Code available upon acceptance.

\begin{table}[t]
\centering
\scriptsize
\setlength{\tabcolsep}{4pt}
\renewcommand{\arraystretch}{1.12}
\caption{Accuracy--efficiency on FG-NET (simulated ground truth). 
$\Delta\text{HC}$: additional human comparisons over EZ-Sort. 
$\tau_{\text{EZ}}$ and $\tau_{\text{Dodger}}$: Kendall's $\tau$ correlation to ground-truth ages. 
EffGain: relative efficiency of each additional comparison vs.\ random labeling (EffGain $>1$ indicates superior information extraction).}
\label{tab:tau_vs_labels}
\begin{tabular*}{0.8\linewidth}{@{\extracolsep{\fill}}ccccc} 
\toprule
\textbf{Size $n$} & \textbf{$\Delta\text{HC}$} & \textbf{$\tau_{\text{EZ}}$} & \textbf{$\tau_{\text{Dodger}}$} & \textbf{EffGain} \\
\midrule
50  & 24 & 0.736 & 0.8122 & \textbf{1.94$\times$} \\
100 & 72 & 0.671 & 0.8440 & \textbf{5.94$\times$} \\
200 & 50 & 0.695 & 0.7935 & \textbf{19.6$\times$} \\
300 & 91 & 0.698 & 0.7469 & \textbf{12.1$\times$} \\
\bottomrule
\end{tabular*}
\end{table}

\begin{table}[t]
\centering
\scriptsize
\setlength{\tabcolsep}{3pt}
\caption{Cross-domain ablation (n=200). We report $\Delta\tau=\tau_{\text{abl}}-\tau_{\text{full}}$ and $\Delta\text{HC}=\text{HC}_{\text{abl}}-\text{HC}_{\text{full}}$ (negative $\Delta\tau$ / positive $\Delta\text{HC}$ indicate worse than full).
\emph{Full ensemble removed} uses Elo only instead of text+Elo+BTL+GP.
\emph{No VLM prior (Elo-IG only)} skips VLM pre-ordering and selects queries via Elo information gain.}
\label{tab:cross_domain_ablation_vertical}
\begin{tabular}{lcccccccc}
\toprule
\multirow{2}{*}{\textbf{Component Removed}} & 
\multicolumn{2}{c}{\textbf{Retina}} & 
\multicolumn{2}{c}{\textbf{Face-Age}} & 
\multicolumn{2}{c}{\textbf{Historical}} & 
\multicolumn{2}{c}{\textbf{Aesthetics}} \\
\cmidrule(lr){2-3}\cmidrule(lr){4-5}\cmidrule(lr){6-7}\cmidrule(lr){8-9}
 & $\Delta\tau$$\downarrow$ & $\Delta\text{HC}$$\uparrow$ & $\Delta\tau$$\downarrow$ & $\Delta\text{HC}$$\uparrow$ & $\Delta\tau$$\downarrow$ & $\Delta\text{HC}$$\uparrow$ & $\Delta\tau$$\downarrow$ & $\Delta\text{HC}$$\uparrow$ \\
\midrule
Hierarchical Prompts & $-0.032$ & $-15.0$ & $-0.030$ & $-27.0$ & $-0.057$ & $-18.0$ & $-0.035$ & $-2.0$ \\
Neural Ranking Head & $+0.243$ & $+388.0$ & $+0.214$ & $+411.0$ & $+0.210$ & $+251.0$ & $+0.173$ & $+394.0$ \\
Full Ensemble & $+0.141$ & $+343.5$ & $+0.183$ & $+360.3$ & $+0.057$ & $+72.7$ & $+0.137$ & $+337.3$ \\
No VLM prior (Elo-IG only) & $-0.333$ & $-111.0$ & $-0.241$ & $-61.0$ & $-0.236$ & $-83.0$ & $-0.380$ & $-63.0$ \\
\midrule
\textbf{Full Dodgersort} & \multicolumn{8}{c}{All components enabled} \\
\bottomrule
\end{tabular}
\end{table}

\subsection{Main Results}

\textbf{Accuracy-efficiency trade-offs.}
Dodgersort achieves superior accuracy efficiency trade-offs across both settings: (1) On human-annotated data (Tables~\ref{tab:inter-rater}--\ref{tab:comparison_count_fgnet}), it matches or improves inter-rater reliability while using equal or fewer comparisons than EZ-Sort (11--16\% reduction across scales), demonstrating higher information extraction per comparison under realistic noise. (2) On simulated ground truth (FG-NET, Table~\ref{tab:tau_vs_labels}), it strategically uses more comparisons to achieve substantially higher accuracy, converting each additional comparison 5--20$\times$ more efficiently than random labeling (EffGain metric), confirming maximum information extraction from human feedback.

\textbf{Inter-rater reliability.}
Table~\ref{tab:inter-rater} shows that pairwise methods substantially outperform classification. Dodgersort consistently achieves highest inter-rater agreement with fewer comparisons: on Retina, Spearman/Pearson of $0.86$ and ICC of $0.95$ with approximately 11\% fewer comparisons than EZ-Sort; on Historical, Spearman/Pearson increase from $0.47$ to $0.60$ and ICC from $0.73$ to $0.82$; on Aesthetics, modest but consistent gains ($Sp{=}0.47$, $Pe{=}0.57$, $ICC{=}0.72$) reflect inherent subjectivity. Table~\ref{tab:comparison_count_fgnet} reports comparison counts averaged across these three datasets, showing consistent efficiency gains across scales: 11\% reduction at $n{=}30$ (80 vs 90 for EZ-Sort), 16\% at $n{=}100$ (400 vs 475).

\textbf{Accuracy-efficiency on FG-NET.}
Table~\ref{tab:tau_vs_labels} examines algorithmic scaling behavior using Face-Age with ground-truth ages as labels, enabling clean comparison with EZ-Sort without confounding from human annotation variability. Dodgersort attains higher Kendall's $\tau$ than EZ-Sort across all scales ($n=50$--$300$). The \textbf{EffGain metric} normalizes accuracy improvement by a label-only upper bound: if each additional comparison fixes at most one discordant pair, then $\Delta\text{HC}$ comparisons improve $\tau$ by at most $2\Delta\text{HC}/P$ where $P{=}\tbinom{n}{2}$. Thus, we define $\text{EffGain} = (\tau_{\text{Dodger}} - \tau_{\text{EZ}}) / (2\,\Delta\text{HC}/P)$. Values consistently $>1$ and growing with $n$ (up to $19.6\times$ at $n{=}200$) indicate that each additional comparison extracts far more information than random relabeling. At $n{=}50$, using slightly more comparisons ($+3\%$) yields substantially higher $\tau$ ($0.812$ vs $0.736$); at $n{=}100$, Dodgersort achieves $\tau{=}0.844$ versus EZ-Sort's $0.671$, demonstrating strategic investment in maximally informative pairs.

\begin{figure}
    \centering
    \includegraphics[width=1.0\linewidth]{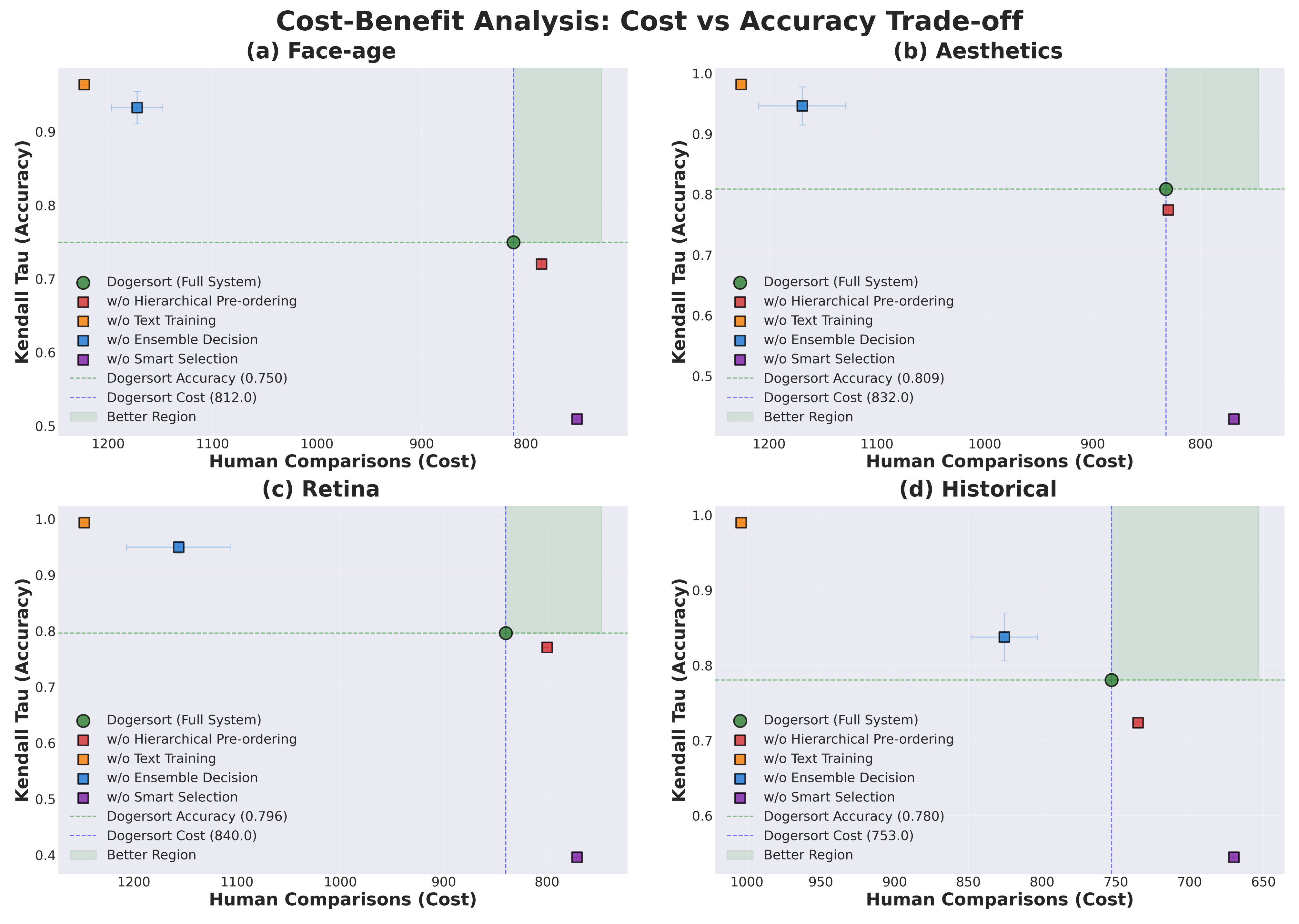}
    \caption{Ablation study across four domains. Full system (green) achieves Pareto-superior trade-offs by balancing accuracy and human comparison cost. Key components (neural ranking head, ensemble, smart selection) each contribute to optimal performance.}
    \label{fig:placeholder}
\end{figure}

\subsection{Ablation Study}

To validate that each component contributes robustly across diverse ranking criteria, we conduct cross-domain ablations at $n=200$ across all four aforementioned datasets, which span both objective and subjective criteria with varying granularity. Table~\ref{tab:cross_domain_ablation_vertical} shows performance differences when removing each component from the full system. Negative $\Delta\tau$ and positive $\Delta\text{HC}$ indicate the full system achieves higher accuracy with fewer human comparisons.

\textbf{Neural ranking head is critical across all domains.} Removing it causes the largest degradation: Kendall's $\tau$ drops by 0.17--0.24 across datasets (e.g., +0.214 on Face-Age means removal lowers $\tau$ by 0.214), and human comparisons increase by 250--411, confirming that learning text-conditioned ranking patterns from feedback is essential regardless of domain characteristics.
\textbf{Full ensemble provides reliable uncertainty.} Using only Elo instead of the full ensemble (text + Elo + BTL + GP) degrades performance substantially: $\tau$ drops by 0.06--0.18 and comparisons increase by 73--361. The ensemble's complementary uncertainty signals enable safer automation decisions.
\textbf{Smart pair selection maximizes information gain.} Removing information-theoretic utility and reverting to random selection shows mixed effects: human comparisons 
decrease by 61--111 (fewer queries), but $\tau$ drops catastrophically by 0.24--0.38 across all domains, confirming that selecting maximally informative pairs is crucial for accuracy-efficiency trade-offs.
\textbf{Hierarchical prompts provide modest but consistent gains.} VLM pre-ordering 
yields smaller improvements ($\Delta\tau$ = 0.03--0.06) at negligible comparison cost (2--27 additional comparisons), demonstrating robust value as a low-cost 
initialization strategy.
Overall, the neural ranking head, ensemble uncertainty, and smart selection are critical for Pareto-optimal performance across diverse subjective and objective ranking tasks, while hierarchical prompts provide reliable initialization.

\section{Discussion and Limitations}
\textbf{Information Efficiency and Robustness.} In human-annotated datasets (Retina, Historical, Aesthetics), where annotation noise reflects realistic expert disagreement, Dodgersort achieves equal or higher inter-rater reliability (Table~\ref{tab:inter-rater}) while using 13\% fewer comparisons on average. This demonstrates effective information extraction under practical noise conditions. In a dataset with simulated ground truth (FG-NET), where synthetic labels eliminate human noise, Dodgersort reveals different strengths. When annotation quality is high, the method strategically invests additional comparisons to achieve substantially higher accuracy: at $n{=}50$, using $+3\%$ more comparisons yields $+10\%$ absolute improvement in Kendall's $\tau$. Critically, the EffGain metric (Table~\ref{tab:tau_vs_labels}) shows that each additional comparison converts into accuracy 5--20$\times$ more efficiently than random relabeling, confirming that these are maximally informative comparisons selected by the ensemble. It suggests that Dodgersort benefits more from high-quality annotations: when annotators are consistent, the framework's uncertainty estimation can exploit this signal more effectively to accelerate convergence.

\textbf{Experimental design.}
The three evaluation paradigms use different datasets by design: inter-rater reliability requires subjective domains where expert noise is realistic (Retina, Historical, Aesthetics), while Face-Age is excluded due to ceiling-level ICC ($\geq$0.97) that masks algorithmic differences. Algorithmic scaling (Table~\ref{tab:tau_vs_labels}) uses Face-Age exclusively because only it provides perception-independent continuous labels at scale---Historical has only 5 coarse decade bins, Aesthetics and Retina lack fine-grained ground truth---ensuring 
discordant pairs are unambiguous for clean $\tau$ measurement.

\textbf{EffGain assumptions.}
The EffGain metric upper-bounds the information value of one comparison by assuming it corrects at most one discordant pair. Under annotation noise, this bound loosens; however, since both Dodgersort and EZ-Sort share identical annotators and noise conditions, EffGain$>1$ constitutes a valid \emph{relative} claim: the additional comparisons chosen by Dodgersort are more informative than random selection, independent of the absolute noise floor.

\textbf{When Dodgersort excels.}
Dodgersort is effective when: (1) annotation budgets are constrained, (2) scale is medium-to-large ($n \ge 50$), (3) criteria have structure enabling hierarchical prompts, and (4) annotator quality is high. In domains with extreme subjectivity (e.g., Aesthetics), gains are more modest but competitive. 

\textbf{Limitations.}(1) For small sets ($n \le 30$), the overhead of adapter training may outweigh the efficiency gains compared to simpler heuristics. (2) While our simulations are robust, human validation was conducted on subsets ($n{=}30$) due to cognitive load and cost constraints; large-scale deployment with domain experts remains for future work. (3) Although asynchronous retraining eliminates annotator idle time, tighter deployments may benefit from replacing the cross-attention head with a lightweight MLP ($<$0.3\,M parameters, ${\approx}1$\,s retraining) or adopting incremental ensemble updates between full retrains.

\section{Conclusion}

Dodgersort integrates VLM pre-ordering, neural ranking heads, ensemble uncertainty estimation, and information-theoretic pair selection for efficient pairwise annotation. 
On human-annotated datasets (Retina, Historical, Aesthetics), we achieve 11--16\% comparison reduction while improving inter-rater reliability (ICC=0.95 vs 0.94 for EZ-Sort). 
The simulation analysis using FG-NET demonstrate a 5--20$\times$ higher information extraction rate (EffGain) per comparison compared to random labeling, which contributes to the overall $\sim$10--16\% improvement in annotation efficiency. 
Cross-domain ablations confirm that all components contribute to Pareto-optimal trade-offs. 
Future work will validate Dodgersort on larger datasets ($n{>}500$) in GP-free mode, explore lightweight dynamic ensemble recalibration, and extend the framework to non-visual ranking domains such as document retrieval.

{\sloppy \subsubsection*{Acknowledgements.} This work was supported by NRF (MSIT) (RS-2024-00338048, RS-2024-00455720, RS-2025-25463760), NIH research projects (2024ER040700, 2025ER040300), High-Performance Computing Support Project (MSIT) (RQT-25-070083), National Supercomputing Center with supercomputing resources (KSC-2024-CRE-0021 \& KSC-2025-CRE-0065), KOCCA (RS-2024-00332210), IITP AI Graduate School (RS-2020-II201373, Hanyang University), and IITP AI Semiconductor Support Program (IITP-(2025)-RS-2023-00253914), and Hankuk University of Foreign Studies Research Fund of 2025.\par}
\subsubsection*{Disclosure of Interests.}
The authors have no competing interests to declare that are relevant to the content of this article.

\bibliographystyle{splncs04}
\bibliography{ref} 
\end{document}